\documentclass[conference]{IEEEtran}
\IEEEoverridecommandlockouts
\usepackage{cite}
\usepackage{hyperref}
\usepackage{amsmath,amssymb,amsfonts}
\usepackage[T1]{fontenc} 
\usepackage{algorithmic}
\usepackage{graphicx}
\usepackage{textcomp}
\usepackage{xcolor}
\usepackage{subfigure}
\usepackage{makecell}
\usepackage{footnote}
\usepackage{url}
\usepackage{lmodern}

\def\BibTeX{{\rm B\kern-.05em{\sc i\kern-.025em b}\kern-.08em
    T\kern-.1667em\lower.7ex\hbox{E}\kern-.125emX}}

\begin{document}

\title{Comparing Feature-based and Context-aware Approaches to PII Generalization Level Prediction}

\author{\IEEEauthorblockN{Kailin Zhang, Xinying Qiu\textsuperscript{\textsection}}
\IEEEauthorblockA{\textit{School of Information Science and Technology} \\
\textit{Guangdong University of Foreign Studies}\\
Guangzhou, China \\
kailinzhang2022@gmail.com, xy.qiu@foxmail.com}
}

\maketitle
\begingroup\renewcommand\thefootnote{\textsection}
\footnotetext{Corresponding author}
\endgroup

\begin{abstract}
Protecting Personal Identifiable Information (PII) in text data is crucial for privacy, but current PII generalization methods face challenges such as uneven data distributions and limited context awareness. To address these issues, we propose two approaches: a feature-based method using machine learning to improve performance on structured inputs, and a novel context-aware framework that considers the broader context and semantic relationships between the original text and generalized candidates. The context-aware approach employs Multilingual-BERT for text representation, functional transformations, and mean squared error scoring to evaluate candidates. Experiments on the WikiReplace dataset demonstrate the effectiveness of both methods, with the context-aware approach outperforming the feature-based one across different scales. This work contributes to advancing PII generalization techniques by highlighting the importance of feature selection, ensemble learning, and incorporating contextual information for better privacy protection in text anonymization.
\end{abstract}

\begin{IEEEkeywords}
personal identifiable information, privacy protection, text anonymization
\end{IEEEkeywords}

\section{Introduction}
The protection of personal privacy, particularly Personal Identifiable Information (PII), is an increasingly critical concern in today's data-driven society. PII refers to any information that can be used to directly identify an individual, such as their name, address, or ID number, as well as quasi-information from which identity can be indirectly inferred through context \cite{1_medlock2006introduction}. Text anonymization aims to safeguard user privacy by identifying and removing or replacing sensitive PII references in documents. Improving PII generalization techniques can have a significant impact on various domains, such as healthcare, finance, and social media, where protecting user privacy is of utmost importance while maintaining data utility for analysis and decision-making.

Current research on text anonymization focuses on three main aspects: 1) PII identification using sequence annotation and named entity recognition methods  \cite{13_szawerna2024detecting,14_silva2020using}; 2) PII de-identification to determine which identified PII need to be concealed based on privacy risk, leveraging concepts like k-anonymity \cite{3_samarati1998protecting} and Privacy-Preserving Data Publishing  \cite{5_dwork2006differential}; and 3) PII generalization, which optimizes the balance between privacy risk and data utility by replacing PII with appropriate, more ambiguous candidates \cite{11_olstad2023generation}.

However, PII generalization faces challenges including 1) uneven data distributions, 2) need for better methodology and 3) imperfect problem definition. Specifically, existing work has assessed model efficacy on structured inputs using large language models such as ELECTRA. The structured inputs include features of PII such as semantic type, while neglecting context information. At the same time, using language models on structured inputs can cause extra computational cost. To address these limitations, this study proposed a feature-based model with an ensemble machine learning algorithm to improve performance on structured model inputs. Furthermore, we also introduced a novel context-aware PII generalization task, constructing the contextual input of the Wikipedia biography dataset \cite{6_papadopoulou2022neural} to incorporate broader information.

Our context-aware approach replaces each PII with candidate generalizations to produce a set of alternative sentences, aiming to select the candidate that best balances privacy and utility. We employed Multilingual-BERT \footnote[1]{https://huggingface.co/google-bert/bert-base-multilingual-uncased} to handle the dataset's language diversity and designed a framework with sentence-level representations, functional transformations, and mean squared error scoring to evaluate candidates. Results demonstrate improved accuracy compared to feature-based models across different dataset scales.

This paper mainly has the following contributions:
\begin{itemize}
    \item Demonstrating that a feature-based machine learning model can outperform the original model on structured inputs from WikiReplace.,
    \item Introducing a context-aware framework and using contextual input that improves generalization performance.
\end{itemize}

Our research aims to pave the way for future research for more effective and efficient PII generalization techniques. We will provide our datasets and codes upon publication.

\section{Related Work}

\subsection{PII Identification and De-identification}
Researchers have proposed PII detection methods in various domains, such as Szawerna et al.'s \cite{13_szawerna2024detecting} LLM-based system for Swedish students' thesis texts and Silva et al.'s \cite{14_silva2020using} named entity recognition combined with machine learning. However, the lack of privacy-oriented annotated resources hinders the evaluation of anonymization methods. Pilán et al. \cite{12_pilan2022text} established the Text Anonymization Benchmark (TAB) to address this issue, which was used by Liu et al.  \cite{15_liu2023deid} and Bubeck et al. \cite{16_bubeck2023sparks} to evaluate their models. To balance privacy risk and data utility, Morris et al.  \cite{2_morris2022unsupervised} used an unsupervised probabilistic model based on k-anonymity, while Papadopoulou et al. \cite{6_papadopoulou2022neural} proposed a hybrid text cleaning method using explicit privacy risk measures and constructed the WikiBioPII dataset for evaluation.

\subsection{PII Generalization}
Olstad\cite{11_olstad2023generation} created the first PII generalization dataset, WikiReplace, using knowledge graphs to generate replacement candidates at different levels for the WikiBioPII dataset. They designed a model using ELECTRA, Categorical MLP, and feature fusion. PII generalization is similar to Lexical Substitution (LS), which replaces target words with semantically similar substitutes. Zhou et al. \cite{17_zhou2019bert} and Qiang et al.\cite{20_qiang2023parals} used BERT and BARTScore/BLEURT for LS, respectively. However, LS methods only consider semantic textual similarity, which may not optimally balance privacy risk and data utility. Our approach addresses this by applying transformations to sentence representations and scoring candidates with MSE.

\section{Methodology}

\begin{figure*}[t!]
   \centering
   \includegraphics[scale=0.6]{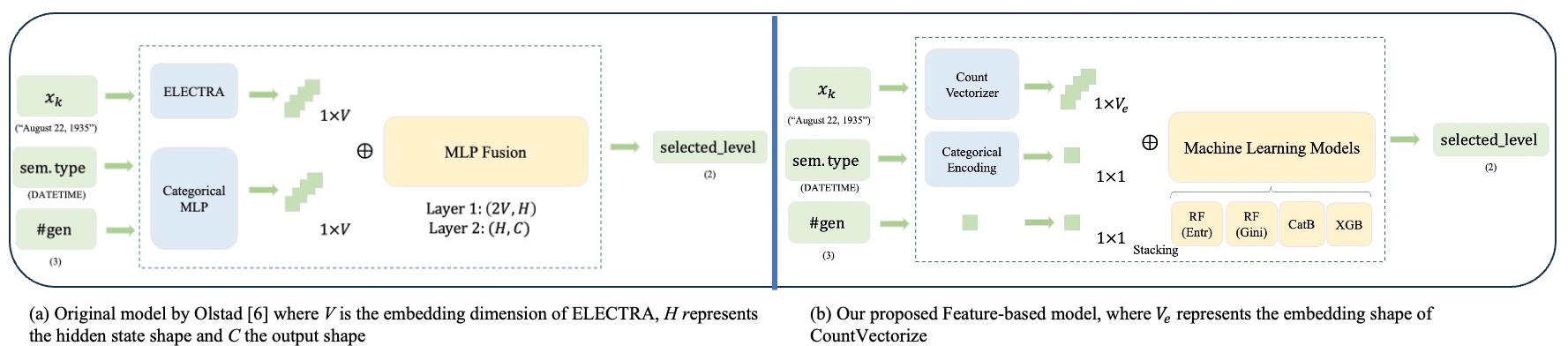}
   \caption{Comparing Original Model and our proposed Feature-based Model}
    \label{Figure 1}
\end{figure*}

\begin{figure*}[t!]
    \centering
    \includegraphics[scale=0.5]{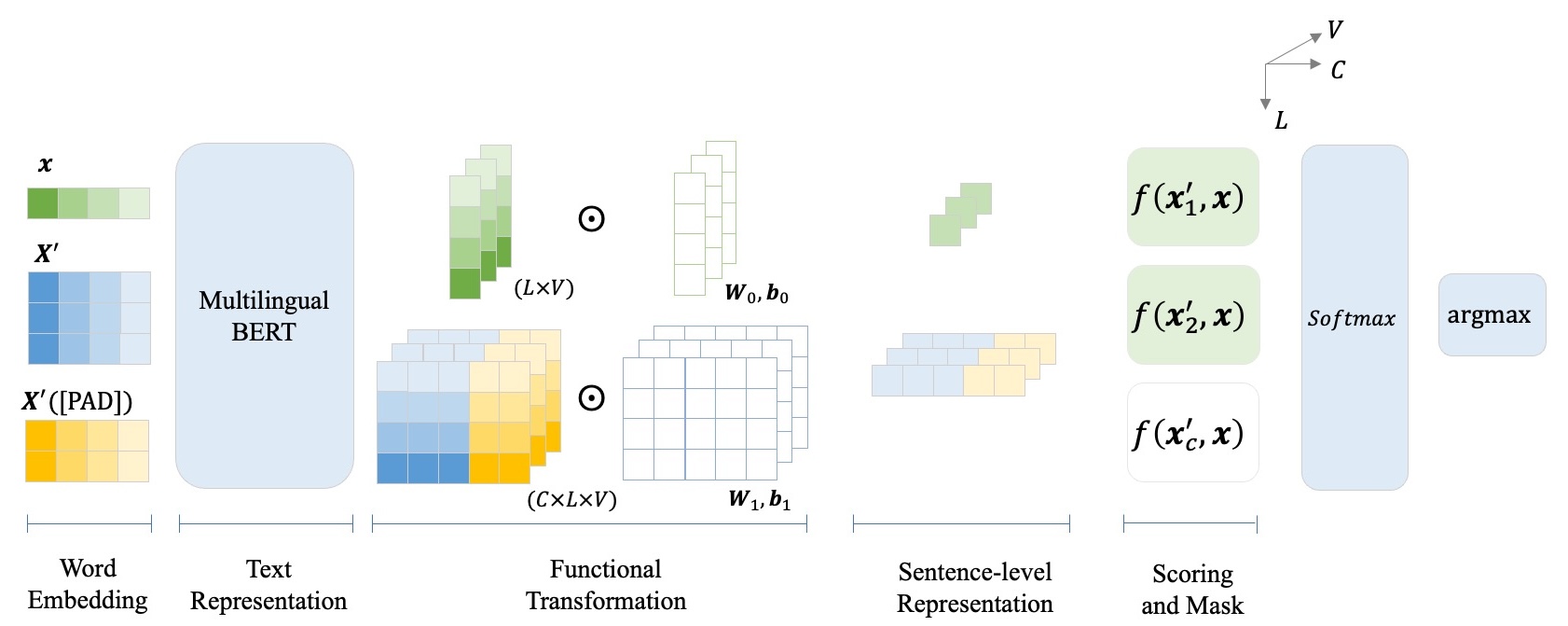}
    \caption{Our proposed Context-aware model}
    \label{Figure 2}
\end{figure*}

\subsection{\textbf{Problem Formulation}}

To clarify the PII prediction task, let's consider the example provided in Table \ref{Table Example}:

\begin{table}[h!]
    \caption{Example of WikiReplace}
    \centering
    \begin{tabular}{|c|c|}
        \hline
        Original Text ($\boldsymbol{x}$) & \makecell[c]{ The person  (born \textbf{August 22, 1935}) \\ is a Canadian lawyer\\ and former Senator. }\\
        \hline
        Text Span (PII) & August 22, 1935 \\
        \hline
        Semantic Type & DATETIME \\
        \hline
        Candidates List & 1935, date in 1930s, *** \\
        \hline
        \# of Generalizations & 3 \\
        \hline
        Selected Generalization & date in 1930s \\
        \hline
        Selected Level & 2 \\
        \hline
        Generalized Text & \makecell[c]{ The person  (born \textbf{[date in 1930s]}) \\ is a Canadian lawyer \\and former Senator. } \\
        \hline
    \end{tabular}
    \label{Table Example}
\end{table}

In this example, the original text $\boldsymbol{x}$ contains a PII "August 22, 1935" that needs to be generalized. The PII has attributes such as semantic type (DATETIME), a list of candidate generalizations ($\boldsymbol{c} =$ [1935, date in 1930s, ***]), the number of generalizations (3), and the selected generalization level (2).

The goal of generalization task is to select the best generalization from a list of suggested replacements for a PII span \cite{11_olstad2023generation}, 

\begin{equation}
    c_{t}^{*} = \mathop{\arg\max}_{c_i \in \boldsymbol{c}}{P(c_{i}|x_k, \boldsymbol{x}, \boldsymbol{c})},
    \label{Equation Original}
\end{equation}

\noindent where $P(\cdot)$ is the probability of selecting candidate $c_i$ predicted by the model, $m$ is the number of candidates in candidates list $\boldsymbol{c}$, $t$ is  selected level, i.e., the position of the chosen generalization in the list of candidates. 

Olstad \cite{11_olstad2023generation} replaces the target variable $c_{t}^{*}$ with the selected level. In the meantime, the researcher only focuses on features (sem.type, \#gen) of PII $x_k$. 
 In other words, the problem is defined as:
\begin{equation}
    {\rm{selected\_level}} = \mathop{\arg\max}_{1 \leq i \leq m}{P(i|x_k, {\rm{sem.type}}, {\rm\#gen.})},
    \label{Equation Feature-based}
\end{equation}
\noindent where $m = {\rm\#gen.}$, i.e., the number of generalizations in $\boldsymbol{c}$.

\subsection{\textbf{Proposed Feature-based Method}}

Considering example in Table \ref{Table Example} and Equation \eqref{Equation Feature-based}, the structured input contains:\\
1) Text Span ($x_k$): "August 22, 1935", \\
2) Semantic Type (sem.type): DATETIME and \\
3) Number of Generalizations (\#gen):2. \\
The output is Selected Level: 2.

As shown in Fig. \ref{Figure 1}(a), Olstad \cite{11_olstad2023generation} designed a classification model. The original PII is represented through ELECTRA. In addition, sem.type and \#gen. are represented through a categorical mlp. The feature representations are then fused with a new-structured MLP. However, this model may have some limitations. The use of ELECTRA for representing short text spans (typically 1-2 tokens) may not be the most efficient choice for the task and the structured input.

As shown in Fig. \ref{Figure 1}(b), we propose a model with machine learning algorithms as follows:
\begin{itemize}
\item Use CountVectorizer\footnote{CountVectorizer: \url{https://scikit-learn.org/stable/modules/generated/sklearn.feature_extraction.text.CountVectorizer.html}} to represent text based on word frequency
\item Represent sem.type through categorical encoding. Since there are 7 categories in PII, we represent each category with a number from 0 to 6. In the meantime, we use \#gen. directly without extra encoding.
\item Apply machine learning algorithms including Random Forest, XGBoost, and CatBoost based on information entropy and Gini coefficient. 
\item Train models with ensemble learning and model stacking to predict selected level of the model.
\end{itemize}

\subsection{\textbf{Proposed Context-aware Method}}

\label {subsection Proposed Context-aware Method}

The structured input from \cite{11_olstad2023generation} ignores the original text $\boldsymbol{x}$ and the candidate list $\boldsymbol{c}$. To incorporate the context information of PII and its candidates, we design a context-aware framework. 

Take Table \ref{Table Example} as example, given the original text $\boldsymbol{x} =$ "The person (born August 22, 1935) is a Canadian lawyer and former Senator." and the candidate list $\boldsymbol{c} =$ ["1935", "date in 1930s", "***"].  We denote  the Selected Generalization ("date in 1930s") as $c_{t}^{*}$, where $t$ is the selected level, i.e., the index of the selected generalization in $\boldsymbol{c}$. Following Equation \eqref{Equation Original}, the \textbf{output} of the context-aware method is $\boldsymbol{x}_t$, the contextualized representation of $c_t$ .\\
{\textbf{(1) Contextual Input Construction}}
\label{subsubsection Contextual Input Construction}

The contextual input of this example is constructed with the following two steps:

\textbf{1. Padding Candidates Set:}
Since PII may have different number of candidates, we firstly pad the candidate list with [PAD], ensuring the length of candidates list for all PIIs to be the same. If we set the maximum number of candidates $C = 5$, the padded list $\boldsymbol{c} =$ ["1935", "date in 1930s", "***", [PAD], [PAD]]. 

\textbf{2. Replacing Original Text:}
We then replace the PII "August 22, 1935" with each candidate to create a list of generalized sentences $\boldsymbol{X}^{\prime}$:

$\boldsymbol{x}_1^{\prime} =$ "The person (born \textbf{[1935]}) is a Canadian lawyer and former Senator."

$\boldsymbol{x}_2^{\prime} =$ "The person (born \textbf{[date in 1930s]}) is a Canadian lawyer and former Senator."

$\boldsymbol{x}_3^{\prime} =$ "The person (born \textbf{[***]}) is a Canadian lawyer and former Senator."

$\boldsymbol{x}_4^{\prime} =$ "The person (born \textbf{[PAD]}) is a Canadian lawyer and former Senator."

$\boldsymbol{x}_5^{\prime} =$ "The person (born \textbf{[PAD]}) is a Canadian lawyer and former Senator."

After the two steps, we use the original text $\boldsymbol{x}$, the replaced text set $\boldsymbol{X}^{\prime}$ as our contextual input.

{\textbf{(2) Model Design}}

As shown in Fig. \ref{Figure 2}, our model consists of the following components.

\textbf{Text Representation:} WikiReplace is a multilingual dataset including English, Chinese, Thai, Hindi, Persian, and other languages. Unlike previous study using ELECTRA, which isn't trained for multilingual data, we propose using Multilingual BERT to learn text representations. We denote $\boldsymbol{h}(\boldsymbol{x})$ and $\boldsymbol{h}(\boldsymbol{X}^{\prime})$ for the original text and the generalized sentences, respectively.

\textbf{Functional Transformation:} We apply a transformation module to the sentence-level representations $\boldsymbol{h}(\boldsymbol{x})$ and $\boldsymbol{h}(\boldsymbol{X}^{\prime})$. In this example, we use a dot product and additive transformation:

\begin{equation}
    \boldsymbol{h}(\boldsymbol{x})_{\text{transformed}}=\boldsymbol{W}_0 \odot \boldsymbol{h}(\boldsymbol{x}) + \boldsymbol{b}_0,
    \label{Equation 3}
\end{equation}

\begin{equation}
    \boldsymbol{h}
    (\boldsymbol{X}')_{\text{transformed}}=\boldsymbol{W}_1\odot \boldsymbol{h}(\boldsymbol{X}')+\boldsymbol{b}_1,
    \label{Equation 4}
\end{equation}

\noindent where $\boldsymbol{W}_0$, $\boldsymbol{b}_0$, $\boldsymbol{W}_1$, and $\boldsymbol{b}_1$ are learnable parameters.

\textbf{Scoring and Mask:} We use the mean squared error (MSE) to measure the information change between the original sentence $\boldsymbol{x}$ and each generalized sentence $\boldsymbol{x}_i^{\prime}$:

\begin{equation}
    f(\boldsymbol{x}_i',\boldsymbol{x})=\dfrac{1}{V}\sum_{v=0}^V(\boldsymbol{h}(\boldsymbol{x}_i')_v-\boldsymbol{h}(\boldsymbol{x})_v)^2,
    \label{Equation 5}
\end{equation}

\noindent where $V$ is the dimension of the sentence-level representations.

To handle cases where the number of candidates is less than the maximum number of generalizations ($C$), we pad the candidate list with [PAD] tokens. After the forward propagation process, we mask the scores of all padded positions to zero. We exemplify the input and output of different methods in Table \ref{Table Dataset Comparison}.

\section{Experiments}
\subsection{\textbf{Dataset}}
\label{subsection Dataset}



\begin{figure}[h!]
    \centering
    \includegraphics[scale=0.5]{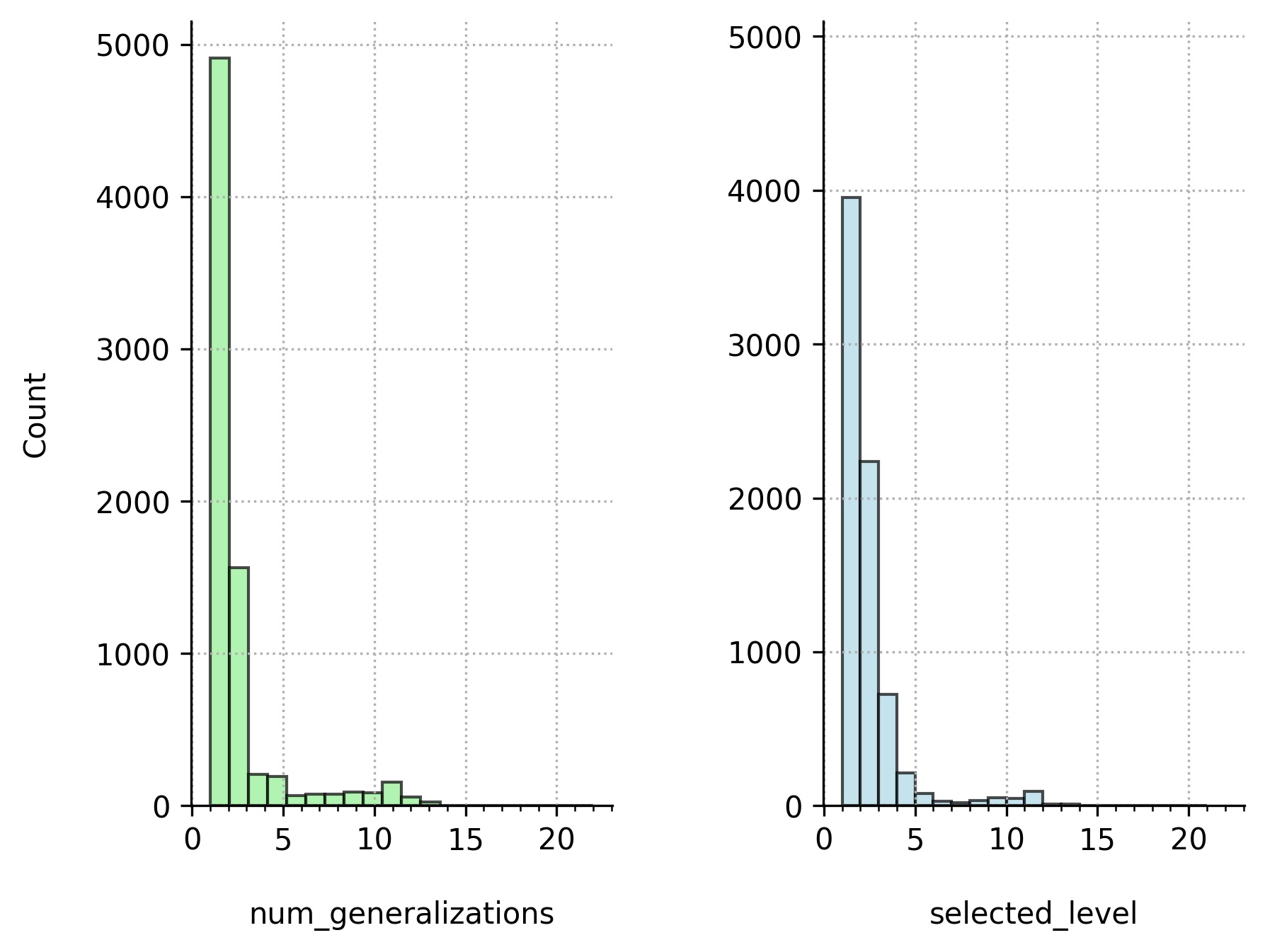}
    \caption{Distribution of number of generalizations and selected level}
    \label{Figure 6}
\end{figure}

Following the input construction methods described in Section \ref{subsubsection Contextual Input Construction}, the different inputs from WikiReplace can be summarized as in Table \ref{Table Dataset Comparison}.  There are 5919 PIIs in the training set and 1587 PIIs in the test set of WikiReplace. The distributions of number of generalizations are shown in Fig. \ref{Figure 6}. We can observe that the distribution is skewed. Most PII have 1-3 generalizations, some have 4-5 generalizations. 
\begin{table*}[h!]
    \centering
    \caption{Comparison of Input and Output of Different Methods }
    \begin{tabular}{|c|c|c|c|}
        \hline
         & Olstad \cite{11_olstad2023generation} (Fig. \ref{Figure 1}(a)) &  Feature-based (Fig. \ref{Figure 1}(b)) & Context-aware Fig. \ref{Figure 2} \\
        \hline
       Input Example & \multicolumn{2}{|c|}{\makecell[c]{"August 22, 1935", \\ DATETIME, 3}} &
       \makecell[c]{ The person  (born \textbf{August 22, 1935})... Senator\\.   ["The person  (born \textbf{[1935]})... Senator.", ...,\\ "The person (born \textbf{[PAD]})... former Senator."] }\\
       \hline
       Output Example & \multicolumn{2}{|c|}{2} & "date in 1930s" \\
       \hline
    \end{tabular}

    \label{Table Dataset Comparison}
\end{table*}

\subsection{\textbf{Evaluation}}

Our restructured contextual input is large in scale, and the distribution of the number of candidates and actual generalization level are uneven. We observe that when $C>7$, the size of partial datasets are closed to the original dataset. We evaluated the performance of the model on different scales with $C=2,5,6,7$ using two metrics,  following Olstad \cite{11_olstad2023generation}:

    \indent\textbf{Majority Vote:} the correct level of generalization is the option that agreed by most annotators.\\
    \indent\textbf{All Selections:} any generalization candidate that is selected by annotators is correct, which is a less strict evaluation metric.

Furthermore, given the inherent imbalance in the dataset’s distribution shown in Fig. \ref{Figure 6}, we calculated precision, recall and F1-scores with a weighted average based on the majority vote as in \eqref{Equation 6}:

\begin{equation}
    {\mathbf{Weighted\,Avg.\,Score} = \sum_{i=1}^C \dfrac{1}{N_i} \mathbf{Score_i}},
    \label{Equation 6}
\end{equation}

\noindent where $N_i$ is the number of data points at generalization level $i$.  We used the leave-one-out validation and early stopping. We provide hyperparameters in Table \ref{Table 6}, and comparison of computational cost in Table \ref{Table cost}.

\begin{table}[h]
    \caption{Implementation Details}
    \centering
    \begin{tabular}{|c|c|}
    \hline
     Parameter    &  Value \\
     \hline
     batch size ($B$) & 2 \\
     \hline
     epoch & 20 \\
     \hline
     max \# of candidates ($C$) & 2, 5, 6, 7 \\
     \hline
     max sequence length ($L$) & 512 \\
     \hline
     embedding dimension ($V$) & 768 \\
     \hline
     loss function & Cross Entropy Loss \\
     \hline
     optimizer & AdamW \\
     \hline
     learning rate & 1e-6 \\
     \hline
     weight decay & 1e-4 \\
     \hline
    \end{tabular}

    \label{Table 6}
\end{table}

\begin{table}[h]
\caption{Computational Costs}
\centering
\begin{tabular}{|c|c|c|c|}
\hline
 & Olstad \cite{11_olstad2023generation} & \makecell[c]{Feature\\based}  & \makecell[c]{Context \\aware} \\ \hline
 Training  (min) & 10$^{\mathrm{a}}$ & 3 & 400$^{\mathrm{b}}$\\ \hline
 Inference  (min) & 0.5 & 0.1 & 0.6\\ \hline
 GPU & 1-Core & none & 1-Core\\ \hline
 \multicolumn{4}{l}{$^{\mathrm{a}}$10 epochs; $^{\mathrm{b}}$20 epochs}
 \end{tabular}
 \label{Table cost}
 \end{table}

\section{Results and Analysis}
\subsection{\textbf{Feature-based Approach Results}}
\subsubsection{\textbf{Main Results}}
As presented in Table \ref{Table 2}, our feature-based machine learning methods outperform both the
baseline and the original approach by Olstad \cite{11_olstad2023generation} when evaluated using the Majority Vote and All Selections metrics.

\label{Feature-based Approach Results}
\begin{table}[h]
\caption{Comparison of Model Accuracy}
\begin{center}
\begin{tabular}{|c|c|c|}
\hline Model & Majority Vote & All Selections \\
\hline
Baseline & 51.36\% & 55.10\% \\
\hline
Olstad \cite{11_olstad2023generation} & 70.29\% & 73.47\% \\
\hline
\bf Feature-based & \bf 77.05\% & \bf 73.63\% \\
\hline
\bf Context-aware $^{\mathrm{a}}$ & \bf 77.81\% & \bf 81.20\% \\
\hline
\multicolumn{3}{l}{$^{\mathrm{a}}$ Evaluated with contextual input when $C = 7$.}
\end{tabular}
\end{center}
\label{Table 2}
\end{table}

\begin{figure}[h!]
    \centering
    \includegraphics[scale=0.5]{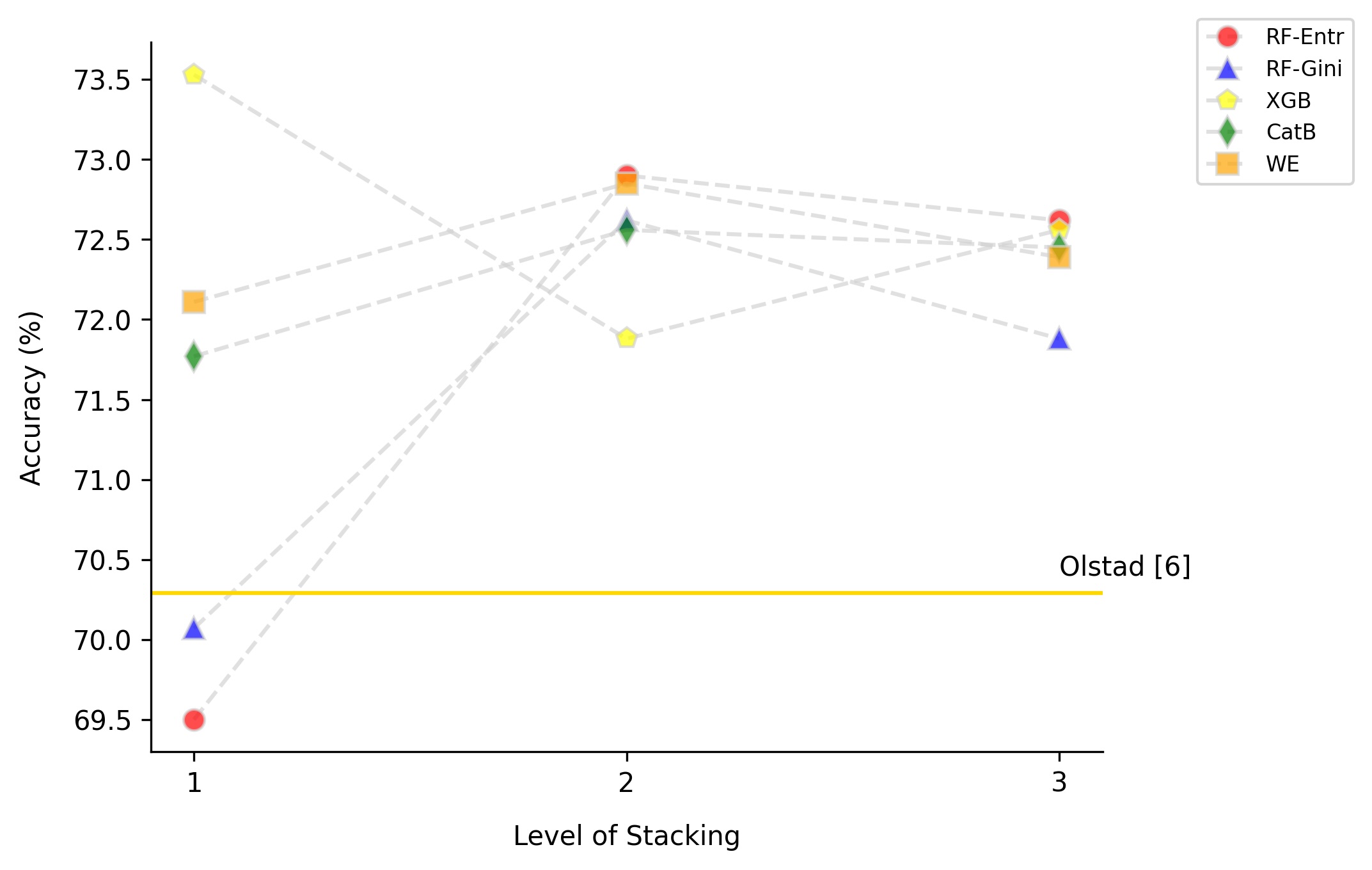}
    \caption{Majority-vote Accuracy of machine learning models}
    \label{Figure 4}
\end{figure} 

\subsubsection{\textbf{Comparing Model Performances}}
Fig. \ref{Figure 4} compares the performances of various machine learning models. XGBoost achieves the highest accuracy at 73.53\% in layer 1, outperforming the original model. These results suggest that for structured inputs, extensive feature engineering and pure machine learning algorithms can yield better performance than deep learning embeddings. Our approach, which focuses on feature selection and ensemble techniques, demonstrates superior accuracy by leveraging the dataset's inherent structure and carefully crafted features. This highlights the importance of considering dataset characteristics when designing PII generalization models and selecting appropriate modeling strategies.
\subsection{\textbf{Context-aware Approach Results}}
\begin{table}[t!]
    \caption{Accuracy with majority vote}
    \centering
    \begin{tabular}{|c|c|c|c|c|c|}
    \hline
     $C$ & Baseline & Olstad &  \makecell[c]{Feature\\based} &  \makecell[c]{Context \\aware} & Dataset \%\\
    \hline
      2 &	67.58\% & 79.21\% & 79.57\% & \bf 81.42\%	& 65.28\% \\
    \hline
      5 &	54.68\%	& 74.20\% & 75.50\% & \bf 75.55\%	& 91.49\% \\
    \hline
      6 &	54.51\%	& 75.31\% & 75.93\% & \bf 77.63\%	& 92.37\% \\
    \hline
      7 &	54.25\%	& 73.11\% & 74.16\% & \bf 77.81\%	& 93.40\% \\
    \hline
    \end{tabular}
    \label{Table 3}
\end{table}

\begin{table}[t!]
    \caption{Accuracy with all selections}
    \centering
    \begin{tabular}{|c|c|c|c|c|c|}
    \hline
    $C$ & Baseline & Olstad &  \makecell[c]{Feature\\based} &  \makecell[c]{Context \\aware} & Dataset \%\\
    \hline
    2 &	70.05\%	& 82.21\% & 82.47\% & \bf 84.22\%   & 65.28\% \\
    \hline
    5 &	58.39\%	& 76.20\% & \bf 79.66\% & 79.01\%	& 91.49\% \\
    \hline
    6 &	58.19\%	& 78.78\% & 79.34\% & \bf 81.62\%	& 92.37\% \\
    \hline
    7 &	57.95\%	& 77.53\% & 78.02\%	& \bf 81.20\%	& 93.40\% \\
    \hline
    \end{tabular}
    \label{Table 4}
\end{table}

\begin{table}[t!]
    \caption{Weighted Average score of context-aware approach}
    \centering
    \begin{tabular}{|c|c|c|c|}
    \hline
    $C$ & \makecell[c]{Weighted \\Avg. Precision} &\makecell[c]{Weighted \\Avg. Recall }& \makecell[c]{Weighted\\ Avg. F1 Score} \\
    \hline
    2 & 81.52\% & 80.20\% & 81.32\% \\
    \hline
    5 &	78.22\%	& 78.07\% & 78.04\% \\
    \hline
    6 &	77.86\%	& 77.63\% & 77.59\% \\
    \hline
    7 &	76.81\%	& 77.50\% &	77.03\% \\
    \hline
    \end{tabular}
    \label{Table 5}
\end{table}

\begin{figure}[h!]
    \subfigure[$C = 5$]{
        \begin{minipage}{.4\linewidth}
            \centering
            \includegraphics[scale=0.5]{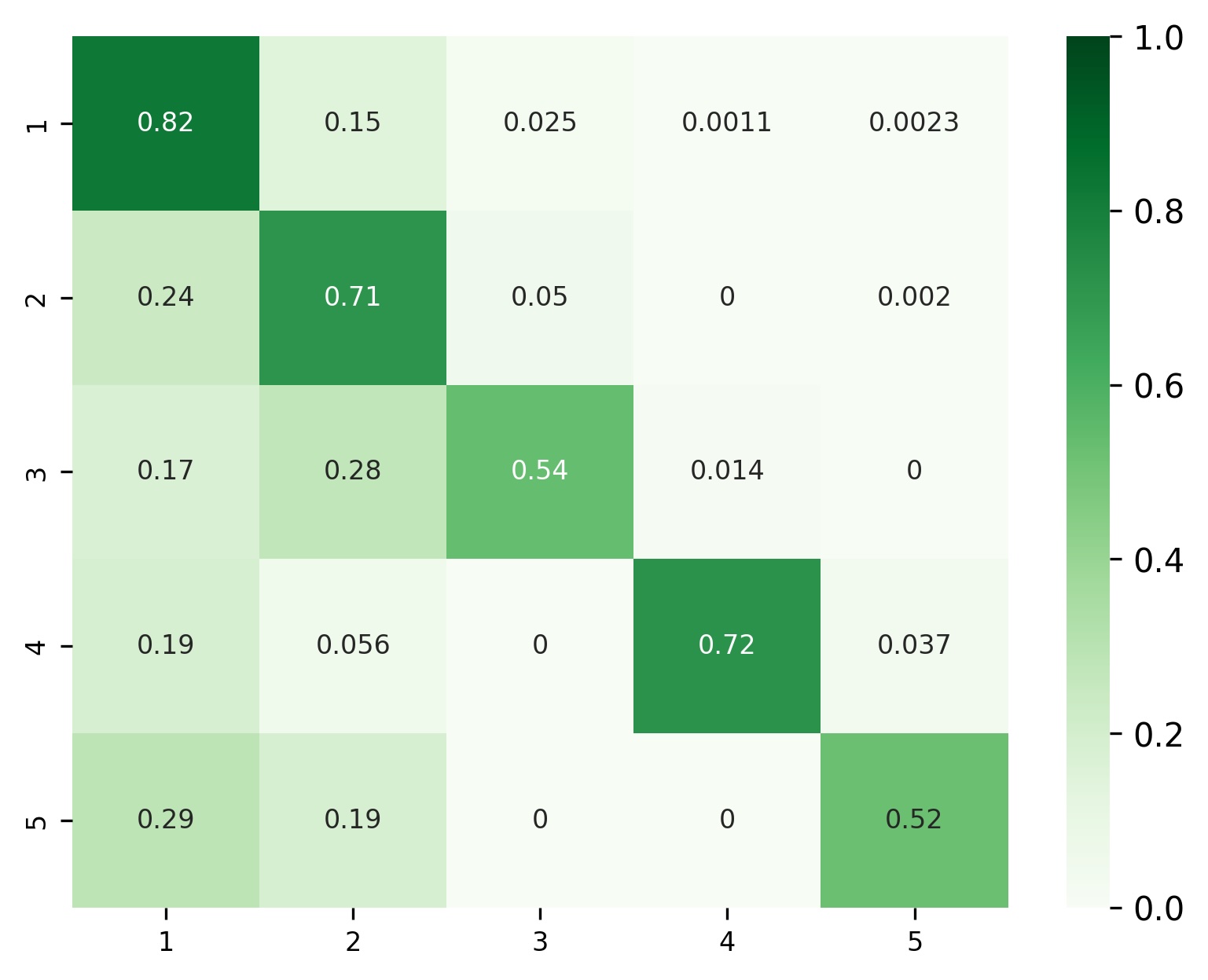}
        \end{minipage}
    } \hspace{-10mm}
    \\
    \subfigure[$C = 6$]{
        \begin{minipage}{.4\linewidth}
           \centering
            \includegraphics[scale=0.5]{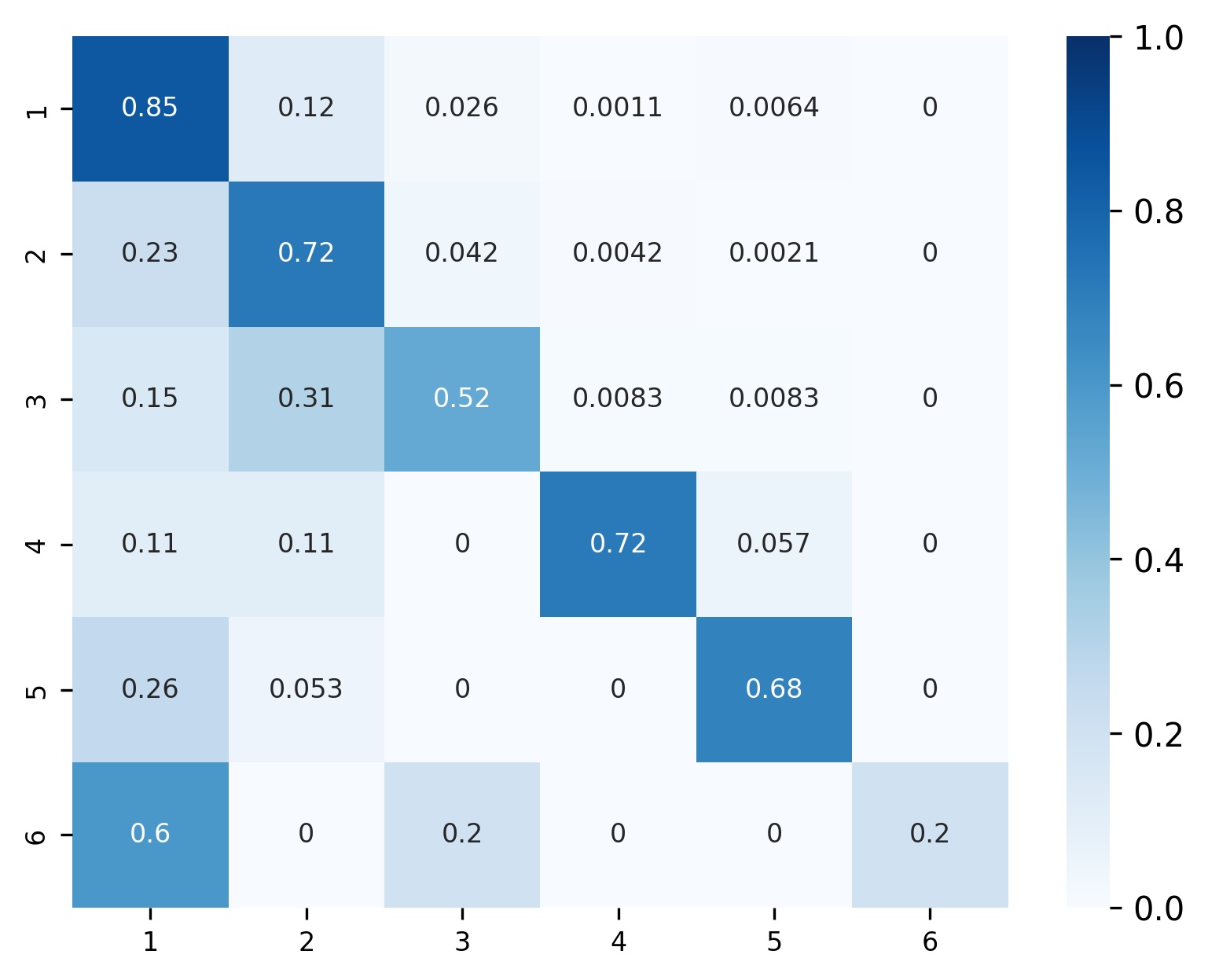}
        \end{minipage}
    } 
    \\
    \subfigure[$C = 7$]{
        \begin{minipage}{.4\linewidth}
           \centering
            \includegraphics[scale=0.5]{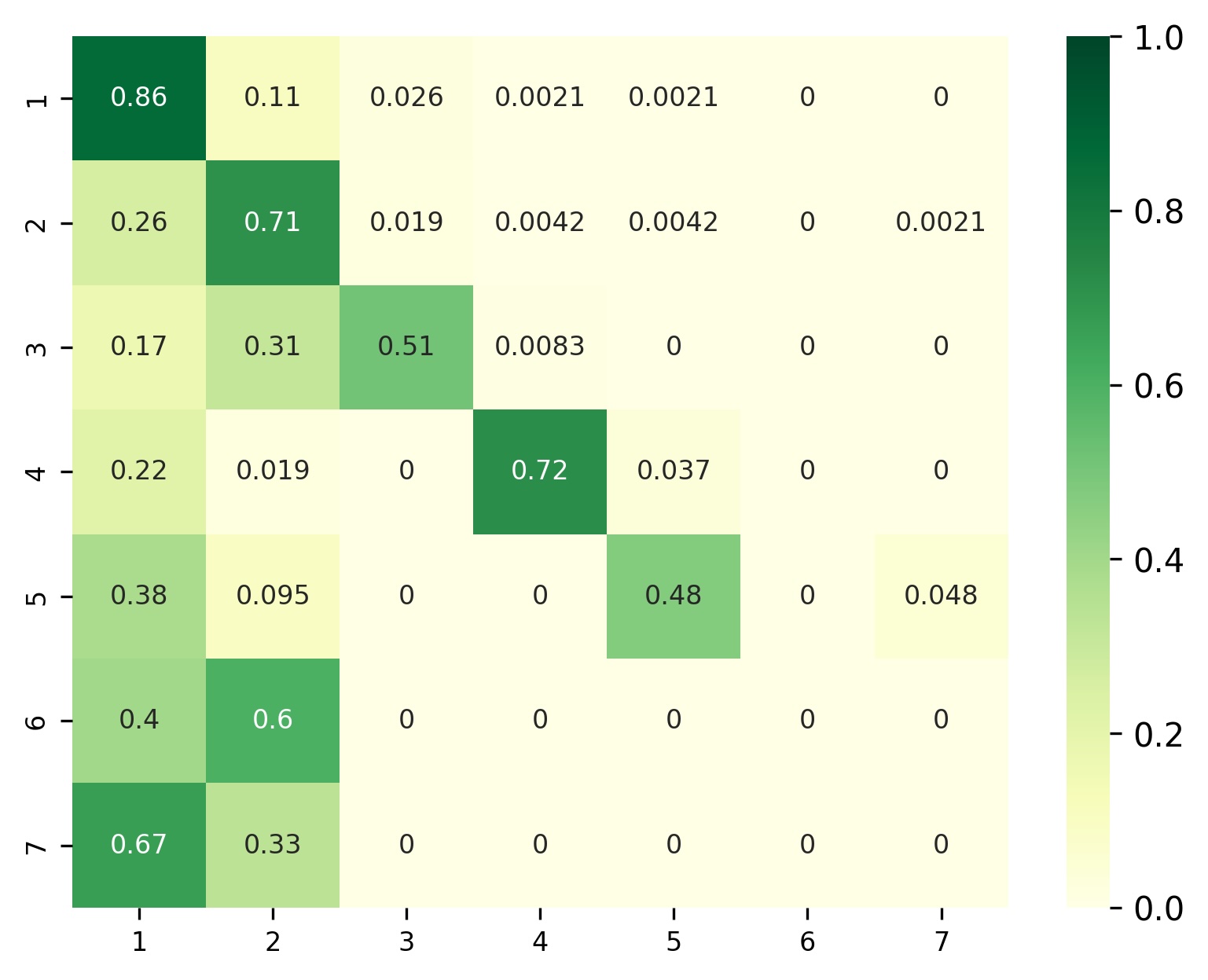}
        \end{minipage}
    }
    \caption{Confusion matrices of Context-aware Method on different scales of dataset, where $C$ is the maximum number of candidates. The numbers in diagonal are the true positive rates.}
    \label{Figure 5}
\end{figure}

\subsubsection{\textbf{Main Results}}
\noindent
The results presented in Table \ref{Table 2} clearly demonstrate the superiority of our context-aware
approach in comparison to the baseline, original, and feature-based methods, which highlights the importance of using contextual information in task.

\subsubsection{\textbf{Ablation Study}}
\begin{itemize}
\item\textbf{Effects of Maximum Number of Candidates}: Tables \ref{Table 3} and \ref{Table 4} demonstrate the context-aware model's consistent performance and robust generalization ability across different dataset scales, even as the task's difficulty increases with the growing number of maximum candidates $C$. Table \ref{Table 5} confirms the absence of significant class imbalance at any dataset scale, further validating the model's reliability and its capacity to perform well across various generalization levels.
\item\textbf{Error Analysis}:
Fig. \ref{Figure 5} presents confusion matrices for the context-aware model when $C = 5, 6, 7$. The model performs well along the diagonal, especially for data points with an actual generalization level of 1, which indicates a good accuracy. However, it struggles with higher generalization levels due to data scarcity. The matrices also reveal the model's tendency to predict lower generalization levels overall and misclassify non-level-1 data points as level 1.
\end{itemize}
\subsection{\textbf{Comparison of Feature-based and Context-aware Approaches}}
Both the feature-based and context-aware approaches outperform the original model. The context-aware method consistently achieving higher accuracy across different dataset scales (Tables \ref{Table 3} and \ref{Table 4}). This superior performance is attributed to the context-aware model's ability to capture contextual and semantic relationships between the original and generalized text, even as the number of candidates increases. The context-aware approach's performance highlights the importance of incorporating contextual information and semantic relationships in PII generalization tasks.

\section{Conclusion}

Our study tackles PII generalization challenges by proposing feature-based and context-aware approaches. The feature-based approach showcases machine learning models' superiority on structured inputs, while the context-aware approach introduces a novel framework that considers contextual relationships, improving accuracy across different dataset scales. Our work advances PII generalization by highlighting feature engineering techniques, and introducing a context-aware framework. These approaches pave the way for future research in developing effective PII generalization techniques to enhance personal privacy protection.

\section*{Limitations}

This study has limitations. Uneven dataset distributions may affect model generalization. Our scoring function for PII generalization could be improved. Computational constraints limited evaluation to 7 candidates. Future research should develop enhanced scoring methods and assess performance on larger candidate sets.

\section*{Acknowledgement}
This work is partially supported by Guangzhou Science and Technology Plan Project (202201010729). We thank the anonymous reviewers for their helpful comments and suggestions.



\bibliographystyle{ieeetr}
\bibliography{citation.bib}

\end{document}